%% file: root.tex
\documentclass[letterpaper, 10 pt, conference]{ieeeconf}  

\IEEEoverridecommandlockouts                              

\usepackage{multirow}
\usepackage{graphicx} 

\usepackage{caption}
\usepackage{subcaption}

\usepackage{url}
\usepackage{xcolor}
\usepackage{algorithm}
\usepackage[noend]{algpseudocode} 

\usepackage{amssymb,amsmath,comment,cite}

\usepackage{amsthm} 

\newtheorem{theorem}{Theorem}

\newtheorem{problem}{Problem}

\usepackage{ifthen}
\newboolean{shortver}
\setboolean{shortver}{true}

\title{\bf Search-Based Path Planning in Interactive Environments among Movable Obstacles}

\author{Zhongqiang Ren$^1$, Bunyod Suvonov$^1$, Guofei Chen$^2$, Botao He$^3$, Yijie Liao$^1$, Cornelia Fermuller$^3$, Ji Zhang$^2$
\thanks{The authors are at $^1$Shanghai Jiao Tong University in China, $^2$Carnegie Mellon University, PA, 15213, and $^3$University of Maryland, MD 20742. Correspondence: zhongqiang.ren@sjtu.edu.cn}
\thanks{This work was supported by the Natural Science Foundation of China under Grant 62403313.}
}

\usepackage{xspace}

\usepackage{etoolbox}
\makeatletter
\patchcmd{\@makecaption}
  {\scshape}
  {}
  {}
  {}
\patchcmd{\@makecaption}
{\\}
{.\ }
{}
{}
\makeatother

\begin{document}

\maketitle

\graphicspath{{./figures/}}

\begin{abstract}
    \input{abstract}
\end{abstract}

\graphicspath{{./figure/}}
 
\section{Introduction}
\input{intro}

\section{Problem Formulation}\label{PPwMO:sec:problem}
\input{problem}

\section{PAMO*}\label{PPwMO:sec:method}
\input{method}

\section{Hybrid-State PAMO*}\label{PPwMO:sec:method2}
\input{method2}

\section{Experimental Results}\label{PPwMO:sec:result}
\input{result}

\section{Conclusion and Future Work}\label{PPwMO:sec:conclude}
\input{conclude}


\bibliographystyle{plain}
\bibliography{references}

\end{document}

%% file: abstract.tex
This paper investigates Path planning Among Movable Obstacles (PAMO), which seeks a minimum cost collision-free path among static obstacles from start to goal while allowing the robot to push away movable obstacles (i.e., objects) along its path when needed. To develop planners that are complete and optimal for PAMO, the planner has to search a giant state space involving both the location of the robot as well as the locations of the objects, which grows exponentially with respect to the number of objects. This paper leverages a simple yet under-explored idea that, only a small fraction of this giant state space needs to be searched during planning as guided by a heuristic, and most of the objects far away from the robot are intact, which thus leads to runtime efficient algorithms. Based on this idea, this paper introduces two PAMO formulations, i.e., bi-objective and resource constrained problems in an occupancy grid, and develops PAMO*, a planning method with completeness and solution optimality guarantees, to solve the two problems. We then further extend PAMO* to hybrid-state PAMO* to plan in continuous spaces with high-fidelity interaction between the robot and the objects. Our results show that, PAMO* can often find optimal solutions within a second in cluttered maps with up to 400 objects.

%% file: intro.tex
Path planning seeks a collision-free path from an initial state to a goal state while avoiding collision among static obstacles, which is of fundamental importance in robotics.
This paper considers a problem called Path planning Among Movable Obstacles (PAMO) where obstacles consist of both movable obstacles (i.e., objects) and static (non-movable) obstacles, and the robot can interact with objects by pushing them away when needed.
PAMO seeks a minimum-cost start-goal path for the robot where both the move and push actions of the robot incur costs, and there is no requirement on the ending poses of the objects.
This problem was shown to be NP-hard~\cite{demaine2000pushpush,wilfong1988motion}, and the challenge is to determine not only a start-goal path among static obstacles but also when and where to interact with the objects.

\begin{figure}[tb]
\centering
\includegraphics[width=\linewidth]{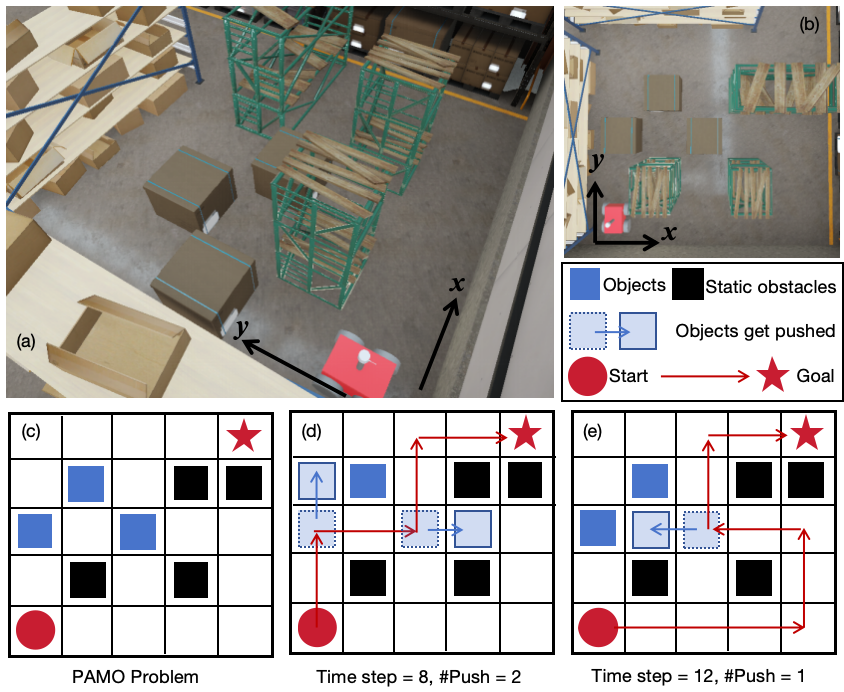}
\caption{PAMO example. (d) and (e) show two alternative solutions, trading off arrival time for number of push. 
    }
\label{PAMO:fig:fig1}
\end{figure}

The goal of this paper is to develop runtime efficient planners for PAMO with completeness and solution optimality guarantees.
For this purpose, we formulate PAMO problems as a search over a grid where robots and obstacles are represented by grid cells, and develop A*-like planners (Fig.~\ref{PAMO:fig:fig1}).
By doing so, the planner has to search a giant state space that includes the location of both the robot and the objects, which thereby grows exponentially with respect to the number of objects.
However, we take the view that, although the state space is huge, in practice, only a small fraction of the state space needs to be explored during planning as guided by a heuristic, and most of the objects that are ``far away'' from the robot are intact, which thus leads to runtime efficient algorithms.

Based on this idea, we first introduce a bi-objective PAMO formulation, which requires minimizing the numbers of both move and push actions simultaneously.
Considering the move and push actions as
two independent dimensions has the potential to avoid naively scalarizing two different types of actions into a single objective.
With multiple objectives, there is often no single solution that optimizes all objectives at the same time, and the problem thus seeks a set of Pareto-optimal solutions, which is usually computationally heavy.
We thus introduce another resource-constrained formulation, which seeks a shortest path while ensuring the number of push actions does not exceed a given limit.
To address the problems, we develop an approach called PAMO*, which leverages our recent work on multi-objective and resource-constrained search~\cite{ren23erca, ren22emoa, 2024_AIJ_EMOA}, and leads to algorithms that can find all Pareto-optimal solutions for the bi-objective PAMO and an optimal solution for the resource-constrained PAMO.
Finally, we seek to get rid of the grid representation by developing Hybrid-state PAMO* (H-PAMO*), which combines PAMO* with hybrid-state A*~\cite{dolgov2010path} and a robot-object interaction simulator based on Box2D~\cite{box2d} to plan in continuous spaces with high-fidelity, at the cost of losing completeness and solution optimality guarantees.

We test PAMO* in various grid maps.
The results show that for the bi-objective problem, our planner can find Pareto-optimal solutions unveiling different ways to interact with the objects, trading off numbers of move and push.
For the resource constrained problem, our planner can often find optimal solutions within a second in cluttered grids with up to 400 objects.
We also verify our H-PAMO* in office-like and warehouse-like environments, and simulate the robot motion in Unity3D, where H-PAMO* plans paths for a robot with kinematic constraints amid rectangular objects that can both translate and rotate.

\subsection{Related Work}
PAMO was formulated in different ways.
When representing the workspace and objects using a grid, PushPush and Push-1 problems are NP-hard~\cite{demaine2000pushpush}.
With polygonal obstacles, navigation among movable obstacles (NAMO) were proposed and shown to be NP-hard~\cite{wilfong1988motion}.
NAMO was studied in environments that are known \cite{stilman2008planning}, partially known~\cite{levihn2014locally}, and unknown~\cite{he2024interactive}, with search-based~\cite{stilman2008planning}, sampling-based~\cite{van2010path}, and learning-based~\cite{xia2020interactive} methods.
This paper differs from these work by considering multi-objective and resource-constrained formulations of PAMO in a fully known grid, and developing planners with completeness and solution optimality guarantees.

Besides, box-pushing games, such as Sokoban, solve puzzles where an agent moves boxes (i.e., objects) from their initial to goal positions in a grid by push.
These games were NP-hard \cite{dor1999sokoban} and has been addressed by heuristic search \cite{junghanns2001sokoban}, Monte-Carlo Tree Search \cite{crippa2022analysis}, Q-learning \cite{wang2006multi}, to name a few.
PAMO differs from these games since it seeks a start-goal path for the robot while these games move the boxes without imposing a goal location on the agent.

Other related work includes manipulation of multiple objects~\cite{9561221,9893496}, objects rearrangement~\cite{krontiris2015dealing}, etc, where various more complicated interaction between the robot and the objects are allowed other than solely push.

Multi-objective search and resource-constrained search are two closely related topics, and a fundamental challenge in these problems is to quickly check paths for dominance, i.e., compare the vector cost of these paths~\cite{hernandez2023simple,ren22emoa,ren23erca,ahmadi2024enhanced}.
This paper leverages these search algorithms~\cite{ren22emoa,ren23erca,2024_AIJ_EMOA} to solve the formulated PAMO problems.

%% file: problem.tex
Let $G=(V,E)$ denote a 2D occupancy grid that represents the workspace, where each cell $v\in V$ has coordinates $v=(x,y) \in \mathbb{Z}^+$ indicating the column and row indices of the cell in the grid.
The time dimension is discretized into time steps.
At any cell $v=(x,y)$, the possible actions of the robot move in one of the four cardinal directions and arrives at one of the neighboring cells $\{(x+1,y),(x-1,y),(x,y+1),(x,y-1)\}$.
Each action takes a time step.
Let $N(v)$ denote the set of neighboring cells of $v\in V$.
Each edge $e=(v_1,v_2) \in E$ indicates the corresponding movement from cell $v_1$ to another cell $v_2$.

At any time, all cells in $V$ are partitioned into three subsets: free space $V_{free}$, static obstacles $V_{so}$, and movable obstacles $V_{mo}$, i.e., $V = V_{free}\cup V_{so} \cup V_{mo}$.
Movable obstacles are also referred to as objects.
Each cell $v\in V_{free}$ is obstacle-free and can be occupied by the robot.
Each cell $v\in V_{so}$ represents a static obstacle and cannot be occupied by the robot at any time.
Each cell $v\in V_{mo}$ indicates an object.
An object $v \in V_{mo}$ can be pushed by the robot to an adjacent cell if all of the following conditions hold: (i) the current cell $v_c$ occupied by the robot is next to $v$; (ii) the robot takes an action to reach $v$; and (iii) the object can be pushed into a neighboring cell $v' = v + (v_c-v)$ that is in free space (i.e., $v' \in V_{free}$).
Note that, Condition (iii) ensures an object cannot be pushed to $v$ if $v$ is occupied by another object.
When the robot is adjacent to an object $v \in V_{mo}$ and moves to cell $v$, the object at $v$ is pushed in the same direction as the robot simultaneously.
In this case, the robot action is counted as a push (as opposed to a move).
Each action, either move or push, takes a time unit.\footnote{The problem formulation and our method can easily adapt to the case where push takes a different amount of time than move. Push may take more fuel than move and is thus considered as a different type of action.}

Let $\pi(v_1,v_\ell) = (v_1,v_2,\cdots,v_\ell)$ denote a path, which is a list of adjacent cells.
Let $\vec{g}(\pi) \in \mathbb{R}^2$ denote the cost vector of $\pi$, where the first element $g_1(\pi)$ is the arrival time at the last cell in $\pi$ (i.e., the total number of move and push), and the second element is the number of push conducted along $\pi$.
Given two vectors $\vec{a}$ and $\vec{b}$ of the same length, $\vec{a}$ \emph{dominates} $\vec{b}$ if every element in $\vec{a}$ is no larger than the corresponding element in $\vec{b}$, and there exists at least one element in $\vec{a}$ that is less than the corresponding element in $\vec{b}$.
Otherwise, $\vec{a}$ is non-dominated by $\vec{b}$.
Given a set of vectors $B$ of the same length, there exists a subset of vectors $B_*$ such that for every vector $\vec{a} \in B$, $\vec{a}$ is non-dominated by any other vector in $B$. Such a set $B_*$ is called the Pareto-optimal front.

Let $v_s$ and $v_g$ denote the start and the goal cells of the robot.
Among all possible paths $\Pi$ from $v_s$ to $v_g$, let $\Pi_* \subseteq \Pi$ denote the {\it Pareto-optimal} set, which is the subset of all paths whose cost vector is within the Pareto-optimal front.
A maximal subset of the Pareto-optimal set, where any two paths in this subset do not have the same cost vector is called a {\it cost-unique} Pareto-optimal set.
This paper considers the following two formulation.

\begin{problem}[BO-PAMO]
    The \underline{B}i-\underline{O}bjective Path \underline{P}lanning \underline{A}mong \underline{M}ovable \underline{O}bstacles (BO-PAMO) seeks a cost-unique Pareto-optimal set (of paths) from $v_s$ to $v_g$.
\end{problem}
    
\begin{problem}[RC-PAMO]
    The \underline{R}esource-\underline{C}onstrained Path \underline{P}lanning \underline{A}mong \underline{M}ovable \underline{O}bstacles (RC-PAMO) seeks a path $\pi$ from $v_s$ to $v_g$ such that $g_2(\pi) \leq K_{push}$, where $K_{push}$ is the maximum number of push the robot can conduct along any path, and $g_1(\pi)$ reaches the minimum.
\end{problem}


%% file: method.tex
\newcommand\algname[1]{\textsf{#1}\xspace}
\newcommand\ERCAstar{\algname{ERCA*}}

\newcommand\procedurename[1]{\textsl{#1}}
\newcommand\IsPrunedByFront{\procedurename{IsPrunByFront}}
\newcommand\IsPrunedByResour{\procedurename{IsPrunByResour}}
\newcommand\FilterAndAddFront{\procedurename{FilterAndAddFront}}
\newcommand\ProcFilter{\procedurename{Filter}}

\newcommand\ResourceLimit{\Vec{r}_{limit}}
\newcommand\NDSymbol{\mathcal{N}\mathcal{D}}
\newcommand\FrontierSet{\mathcal{F}}
\newcommand\DomSymbol{\preceq}
\newcommand\NonDomSymbol{\npreceq}
\newcommand\LexLessSymbol{<_{lex}}
\newcommand\LexLargerSymbol{>_{lex}}


\subsection{PAMO* Search}
PAMO is the problem name while PAMO* (with *) is the method name.
Let $m=|V_{mo}|$ denote the number of objects, and let position vector $p = (v_1,v_2,\cdots,v_m)$ denote the cells occupied by all objects.
Let $s=(v, p)$ denote a search state, where $v \in V_{free}$ is the cell occupied by the robot, and $p$ is an aforementioned position vector.
In other words, the state space of the search is $S=G\times G \times \cdots \times G=G^{m+1}$, which encodes the locations of both the robot and all objects.

Let $s_o$ denote the initial state, where the robot is at $v_s$ and all objects are at their original positions.
Different from conventional A*, where the search only needs to record one optimal path $\pi$ (via parent pointers) from $s_o$ to any state $s$, PAMO* have to store multiple non-dominated paths from $s_o$ to $s$ and differentiate between these paths.
To do so, let $l = (s,\vec{g})$ denote a label, where $s$ is a state as aforementioned and $\vec{g}$ is the cost vector of a path from $s_o$ to $s$.
Labels are compared based on their cost vectors and two labels are non-dominated by each other if their cost vectors are non-dominated by each other.
We use $s(l),\vec{g}(l)$ to denote the state and the cost vector contained in $l$ respectively. We use $v(l)$ to denote the cell occupied by the robot in state $s(l)$.

Let $\FrontierSet(s)$ denote the \emph{frontier set} at state $s$, which is a set of labels, where any pair of labels in $\FrontierSet(s)$ are non-dominated by each other.
Similar to A*, let $\vec{g}(s)$ denote the cost-to-come, i.e., the cost vector of the path from $s_o$ to $s$ and let $\vec{h}(s)$ denote the heuristic vector of state $s$ that estimates the cost-to-go.
Let $\vec{f}(s):=\vec{g}(s) + \vec{h}(s)$ denote the $f$-vector of state $s$, and let $\mathcal{O}$ denote an open list of states, which is a priority queue that prioritize states based on their $f$-vectors from the minimum to maximum in lexicographic order.
Let $L^*$ denote the set of labels representing solution paths that are found during the search.

PAMO* (Alg.~\ref{PPwMO:alg:PAMO}) begins by creating the initial label $l_o = (s_o, \vec{g}=\vec{0})$, which is added to $\mathcal{O}$.
The frontier set is initialized as an empty set for any state.
In a search iteration (Line~\ref{alg:line:PAMO_while_begin}-\ref{alg:line:PAMO_while_end}), a label $l$ with the lexicographically minimum $f$-vector in $\mathcal{O}$ is popped, and is then checked for pruning (Line~\ref{alg:line:PAMO_check1}), which is elaborated later.
If $l$ is not pruned, $l$ is used to update the frontier set $\FrontierSet(s)$, where $l$ is compared against any existing label $l' \in \FrontierSet(s)$.
If $l'$ is dominated by $l$, $l'$ is removed from $\FrontierSet(s)$.
Then, $l$ is added to $\FrontierSet(s)$.
As a result, $\FrontierSet(s)$ always contains non-dominated labels.

After updating the frontier, PAMO* checks if $l$ reaches the goal (i.e., $v(l) = v_g$).
When solving RC-PAMO where only one optimal path is required, PAMO* terminates if $v(l) = v_g$ and the path corresponding to label $l$ is guaranteed to be optimal.
When solving MO-PAMO where all cost-unique Pareto-optimal paths are required, PAMO* skips the rest of the current search iteration and continues with the next search iteration if $v(l) = v_g$.
When solving MO-PAMO, PAMO* terminates when $\mathcal{O}$ depletes, i.e., all states in $\mathcal{O}$ are popped and are either expanded or pruned.

If label $l$ does not reach the goal, then $l$ is expanded by finding successor states of $s(l)$ (Line~\ref{alg:line:PAMO_for_begin}).
Specifically, the procedure GetSuccessor returns the set of all reachable states from the given state $s(l)$, where each of the successor states update both the cell of the robot, and the cells of the objects when any object is pushed by the robot.
Then, for each of the successor state $s'$, a corresponding label $l'=(s',\vec{g}')$ is created, where $\vec{g}' = \vec{g}(l) + \text{GetCost}(s,s')$ and the GetCost function returns either a cost vector $(1,0)$ if the robot moves without pushing any object, or $(1,1)$ if the robot moves while pushing an object.
Afterwards, the new label $l'$ is checked for pruning (Line~\ref{alg:line:PAMO_check2}).
If $l'$ is not pruned, the $f$-vector and the parent pointer related to $l'$ are updated and $l'$ is added to $\mathcal{O}$ for future search.

At the end of the search, each label in $L^*$ represents a solution path, which is reconstructed by iteratively tracking the parent pointers, and the solution path(s) are returned.

\begin{algorithm}[tbp]
	\caption{PAMO*}\label{PPwMO:alg:PAMO}
 \small
	\begin{algorithmic}[1]
		\State{$l_o \gets (s_o, \vec{0})$, $\Vec{f}(l_o) \gets \Vec{0} + \Vec{h}(s_o)$}\label{alg:line:PAMO_init_label}
		\State{Add $l_o$ to OPEN}
		\State{$\FrontierSet(s)\gets \emptyset, \forall s \in S$, $L^* \gets \emptyset$}
		\While{OPEN $\neq \emptyset$}\label{alg:line:PAMO_while_begin}
		\State{$l \gets $ OPEN.pop()}
		\If{\textit{FrontierCheck}($l$) \textbf{or} \textit{SolutionCheck}($l$)}\label{alg:line:PAMO_check1}
		\State{\textbf{continue}}\Comment{Current iteration ends}
		\EndIf
		\State{\textit{UpdateFrontier}($l$)}\label{alg:line:PAMO_updateFront}
		\If{\textit{ReachGoal}($l$)}
		\State{Add $l$ to $L^*$}
		\State{\textbf{break}}\Comment{RC-PAMO*}
		\State{(or \textbf{continue})}\Comment{MO-PAMO*}
		\EndIf
		\ForAll{$s' \in$ GetSuccessors($s(l)$)}\label{alg:line:PAMO_for_begin}
		\State{$l' \gets (s', \vec{g}(l) + \vec{c}(s,s'))$}\label{alg:line:PAMO_gen_label}
		\If{\textit{FrontierCheck}($l'$) \textbf{or} \textit{SolutionCheck}($l'$)}\label{alg:line:PAMO_check2}
		\State{\textbf{continue}}\Comment{Move to the next successor.}
		\EndIf
		\State{$\vec{f}(l') \gets \vec{g}(l') + \vec{h}(s(l'))$, $parent(l')\gets l$}\label{alg:line:PAMO_f_vec_calc}
		\State{Add $l'$ to OPEN}
		\EndFor
		\EndWhile\label{alg:line:PAMO_while_end}
		\State{\textbf{return} Reconstruct($L^*$)}
	\end{algorithmic}
\end{algorithm}

\subsection{Procedures in PAMO*}
\subsubsection{GetSuccessors}
The procedure FrontierCheck in PAMO* takes a state $s=(v,p)$ and returns its successor states.
It considers all possible neighboring cells of the robot $N(v)$, and for each of them, there are three cases.
First, the robot moves to a free cell, which yields a successor state $s'=(v',p)$ where $(v.v') \in E$ and the position vector $p$ remains the same as in $s$.
Second, the robot moves to a cell that is a static obstacle, which is not a valid move and yields no successor.
Third, the robot moves to a cell that is occupied by an object $p_k$ (the $k$-th element in the position vector $p$).
In this case, the procedure further predicts the motion of the object $p_k$ and checks if the object can be pushed to an adjacent free cell $u$. If so, a successor state $s'=(v',p')$ is generated, where $(v,v') \in E$ and the position vector $p'$ is updated by first copying $p$ and then modifying $p_k$ to be $u$.
Otherwise (i.e., $p_k$ is pushed to a static obstacle or another object), no successor state is generated.

\subsubsection{Heuristic Computation}
PAMO* calculates the heuristic vectors as follows. PAMO* first runs a pre-processing before the search starts, which invokes a Dijkstra search backwards from $v_g$ to all other vertices in the grid while ignoring any objects.
By doing so, for each cell $v\in V_{free}$, we know the distance $d^*(v)$ to $v_g$ along a shortest path among static obstacles.
We then use $(d^*(v),0)$ as the heuristic vector for any label $l$ whose robot position is at $v$, ignoring object positions.
This heuristic is a lower bound on the true cost-to-go from $v$ to $v_g$ for the number of both move and push.

\subsubsection{FrontierCheck}
The procedure FrontierCheck in PAMO* is the same when solving both RC-PAMO and MO-PAMO, where a given label $l$ is checked against any existing label $l' \in \FrontierSet(s(l))$ for dominance.
If the cost vector $\vec{g}(l')$ of any existing label $l' \in \FrontierSet(s(l))$ is component-wise no larger than $\vec{g}(l)$, then $l$ cannot lead to an optimal path and should be pruned, and FrontierCheck returns true.
Otherwise, FrontierCheck returns false.

\subsubsection{SolutionCheck}
The procedure SolutionCheck is different when solving RC-PAMO and MO-PAMO.
When solving MO-PAMO, SolutionCheck compares the $f$-vector of the given label $l$ against the $g$-vector of any existing label $l' \in \mathcal{F}(v_g)$.
Note that each $l' \in \mathcal{F}(v_g)$ represents a solution path from $v_s$ to $v_g$ that is already found during the search, and $\vec{g}(l')=\vec{f}(l')$ since $\vec{h}(l') = 0$.
If $\vec{g}(l')$ is component-wise no larger than $\vec{g}(l)$, then $l$ cannot be a Pareto-optimal solution and should be pruned, and SolutionCheck returns true.
Otherwise, SolutionCheck returns false.

When solving RC-PAMO, SolutionCheck compares $f_2(l)$, the second component of the $f$-vector of the given label $l$, against $K_{push}$ the limit on the number of push actions.
If the limit is exceeded ($f_2(l) > K_{push}$), then $l$ cannot be a solution and is thus pruned, and SolutionCheck returns true.
Otherwise, SolutionCheck returns false.

\subsection{Discussion}
\subsubsection{Implicit State Generation}
The state space $S$ is never created explicitly, i.e., allocate the memory for each state before the search starts.
Instead, the state space is created implicitly, i.e., the states and the frontier sets are created only when the search generates the states.
PAMO* only requires a GetSuccessors procedure to create successors out of a given state in $S$, and never requires the full knowledge of $S$.

\subsubsection{Giant State Space}
PAMO* has a small constant branching factor, which is the number of successors returned by GetSuccessors procedure.
Although the states space $S$ is extremely large and grows exponentially with respect to the number of objects.
In practice, guided by the heuristic, PAMO* often needs to explore only a small fraction of $S$ before finding the (Pareto-)optimal solution(s), even if there are many objects.
Intuitively, most of the objects that are far away from the robot's path from $v_s$ to $v_g$ are never touched, and the corresponding element of them in the position vector $p$ are never changed during the search.

\subsubsection{Global and Local Checks}
Intuitively, FrontierCheck can be regarded as a ``local'' check which compares a label against the existing non-dominated label at the same state.
Correspondingly, SolutionCheck can be regarded as a global check which compares a label against ``global'' information, i.e., either the existing solution paths that have been found or the resource limit.

\subsubsection{Action Costs}
In the problem formulation, both move and push take a time unit.
Our method PAMO* can easily handle the case where push and move take different amount of time, or more generally speaking, incur various types of costs, as long as the costs can still be described by cost vectors and are still additive.
The only place to be modified is Line~\ref{alg:line:PAMO_gen_label} in Alg.~\ref{PPwMO:alg:PAMO} when calculating $\vec{c}(s,s')$ between two states $s$ and $s'$.

\subsection{Properties}
Let RC-PAMO* (and MO-PAMO*) denote the version of PAMO* when using Alg.~\ref{PPwMO:alg:PAMO} to solve RC-PAMO problem (and MO-PAMO problem, respectively).
The completeness and solution optimality of PAMO* are inherited from EMOA*~\cite{ren22emoa} and ERCA*~\cite{ren23erca}.
In particular, RC-PAMO* (and MO-PAMO*) can be regarded as applying ERCA* (and EMOA*) onto the new state space $S$ with the new way of successor generation.
\begin{theorem}
MO-PAMO* is complete and can find all cost-unique Pareto-optimal solutions for MO-PAMO.
\end{theorem}
\begin{theorem}
RC-PAMO* is complete and can find an optimal solution for RC-PAMO.
\end{theorem}

%% file: method2.tex

Based on PAMO*, we further develop Hybrid-state PAMO* (H-PAMO*) by leveraging the idea in hybrid-state A*~\cite{dolgov2010path} to plan the robot motion in continuous space and time, handle kinematic constraints of robots, and consider more detailed interaction between the robot and the objects.
Same as Hybrid-state A* for path planning in continuous space, our H-PAMO* loses completeness and solution optimality guarantees for PAMO in continuous space.

\subsection{Environment and Robot}
We consider a first order unicycle model, where the robot pose is $\xi=(x,y,\theta) \in SE(2)$ and control is $u=(v,\omega) \in \mathcal{U}$ where $v$ is the linear velocity and $\omega$ is the angular velocity, and $(v,\omega)$ are subject to control limits.
The robot satisfies the system dynamics $\dot{\xi}=(\dot{x},\dot{y},\dot{\theta}) = f_{dyn}(\xi,u) = (v\sin(\theta),v\cos(\theta),\omega)$.
The workspace $\mathcal{W}$ is a bounded 2D Euclidean space with a set of static rectangle obstacles $\mathcal{W}_{obs}$.
In the free space $\mathcal{W}_{free} = \mathcal{W}\backslash\mathcal{W}_{obs}$, there are rectangle objects.
A search state now includes both the pose of the robot and the poses of all objects.

The robot-object interaction is described by a black-box function $f_{sim}$ which takes (i) a state (i.e., the poses of the robot and the objects), (iii) a control of the robot, and (iv) a small amount of time $dt$, conducts a forward simulation and returns the ending poses of the robot and the objects.
Here, $f_{sim}$ only simulates the poses of the robot and the objects, and ignores the higher order terms (e.g. velocities).

\subsection{H-PAMO* Search}
H-PAMO* is similar to PAMO* with the following differences.
\textbf{First}, to simplify the presentation, H-PAMO* is developed as a single-objective algorithm, where all actions (either move or push) incur an action time, and H-PAMO* minimizes the arrival time at the goal.
As a result, all cost vectors $\vec{g},\vec{h},\vec{f}$ (Lines~\ref{alg:line:PAMO_init_label}, \ref{alg:line:PAMO_gen_label}, \ref{alg:line:PAMO_f_vec_calc}) becomes scalar values and the transition cost $\vec{c}$ is now a scalar representing the action time from one state to another.
\textbf{Second}, H-PAMO* plans in a continuous space.
To compare and prune paths (FrontierCheck on Lines~\ref{alg:line:PAMO_check1}, \ref{alg:line:PAMO_check2} and UpdateFrontier on Line~\ref{alg:line:PAMO_updateFront}), H-PAMO* uniformly discretizes the workspace into mutually exclusive cells of size $\Delta x \times \Delta y \times \Delta\theta$, and every pose belongs to a unique cell.
The state space $S$ is thus discretized into a grid $S_g$, where each cell is of size $(\Delta x \times \Delta y \times \Delta\theta)^{|V_{mo}|+1}$.
When two paths end with states that belong to the same cell in $S_g$, H-PAMO* only stores the cheaper path and prunes the other.
Here, $S_g$ is never created explicitly and any cell in $S_g$ is only created when a state in that cell is generated.
\textbf{Third}, since H-PAMO* optimizes the arrival time, and there is no limit on the number of push actions, there is no SolutionCheck on Line~\ref{alg:line:PAMO_check1}, \ref{alg:line:PAMO_check2} in H-PAMO*.
\textbf{Fourth}, to get successors, H-PAMO* considers a finite set of motion primitives that are generated by sampling controls from the control space $\mathcal{U}$ and running forward simulation for each of the sampled control for a short amount of time $dt$.
When generating a successor, the interaction between the robot and the objects needs to be considered. In our implementation, we use Runge-Kutta 4th order method to integrate the $f_{dyn}$ to simulate the robot motion given the control, and use Box2D~\cite{box2d} to implement $f_{sim}$ to predict the motion of the objects when pushed by the robot.
\textbf{Finally}, H-PAMO* terminates when the search finds a path that reaches the goal pose within a given tolerance $\epsilon_{goal} \in \mathbb{R}^3$.

%% file: result.tex
We test PAMO* in various grid maps from a public dataset~\cite{gridDataset}.
Each grid map has static obstacles of different densities.
We place objects $V_{mo}$ randomly in the grid without overlapping with static obstacles or start and goal positions.
We create 10 instances for each map, and the instances are not guaranteed to be feasible due to the randomly located objects.
We set a 1 minute runtime limit for each instance.
All tests are conducted on a MacBook laptop with a M2 Pro CPU and 16GB RAM.

We first test MO-PAMO* and RC-PAMO* (with $K_{push} = \infty$) in a (fixed) 8x8 empty grid map with varying percentage of movable obstacles 10\% ($|V_{mo}| = 6$), 20\% ($|V_{mo}| = 12$), and 30\% ($|V_{mo}| = 19$).
We then fix the percentage of movable obstacles to 10\% and change the maps to a Random 32x32 ($|V_{mo}| = 102$), Room 32x32 ($|V_{mo}| = 102$), and Random 64x64 ($|V_{mo}| = 409$).
We report the runtime and number of expansions (i.e., Alg.~\ref{PPwMO:alg:PAMO} reaches Line~\ref{alg:line:PAMO_for_begin}).

\begin{figure}[tb]
\centering
\includegraphics[width=\linewidth]{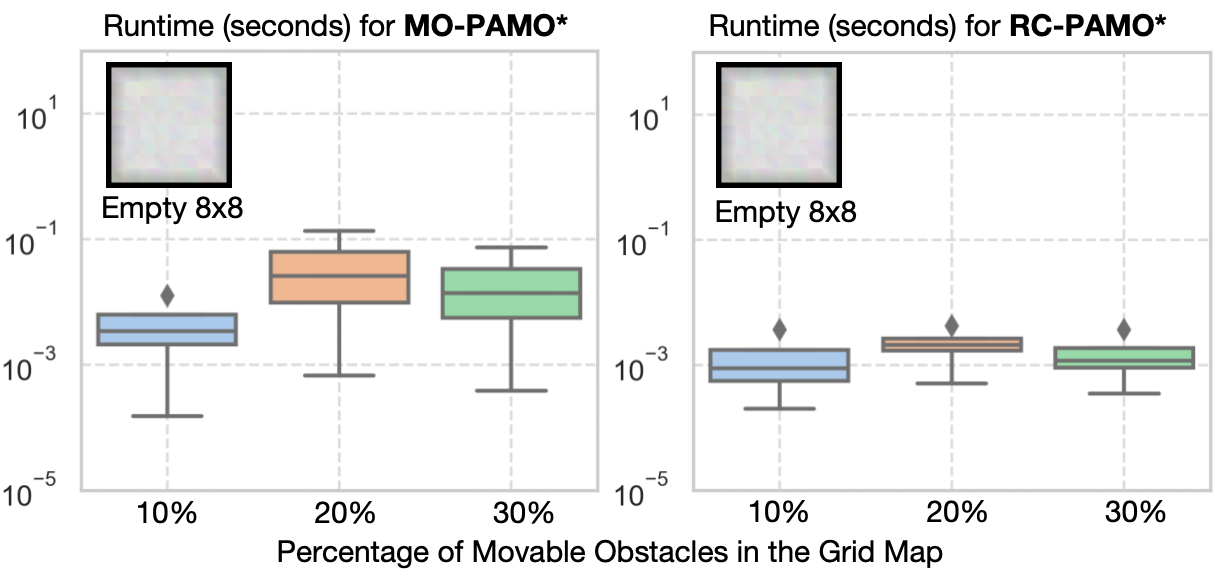}
\caption{Runtime of MO-PAMO* and RC-PAMO* in Empty 8x8 grid map with varying percentage of objects.
    }
    \vspace{-2mm}
\label{PAMO:fig:res_rt_empty}
\end{figure}

\begin{figure}[tb]
\centering
\includegraphics[width=\linewidth]{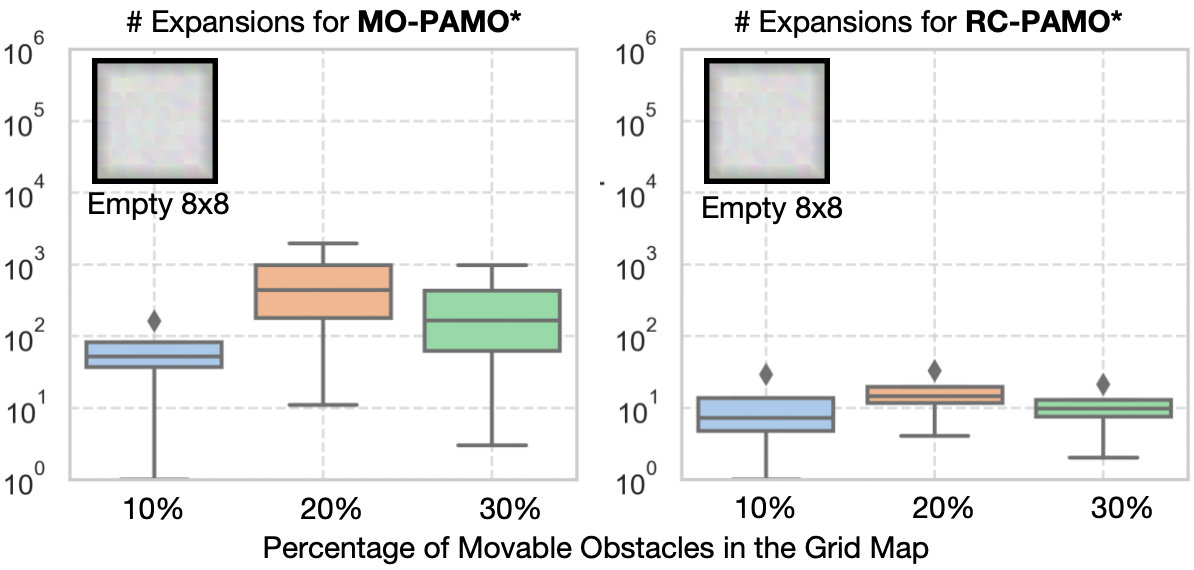}
\caption{Numbers of expansion of MO-PAMO* and RC-PAMO* in the Empty map with varying object percentage.
    }
    \vspace{-4mm}
\label{PAMO:fig:res_exp_empty}
\end{figure}

\begin{figure}[tb]
\centering
\includegraphics[width=\linewidth]{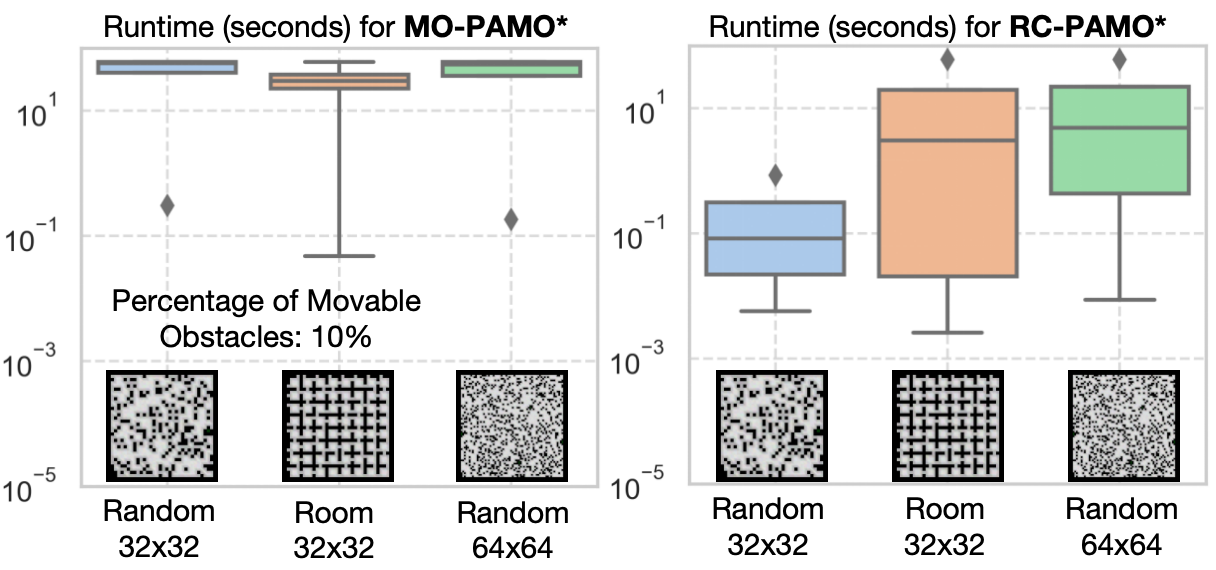}
\caption{Runtime of MO-PAMO* and RC-PAMO* in three grid maps Random 32x32, Room 32x32 and Random 64x64, with 10\% objects.
    }
\label{PAMO:fig:res_rt_random}
\end{figure}

\begin{figure}[tb]
\centering
\includegraphics[width=\linewidth]{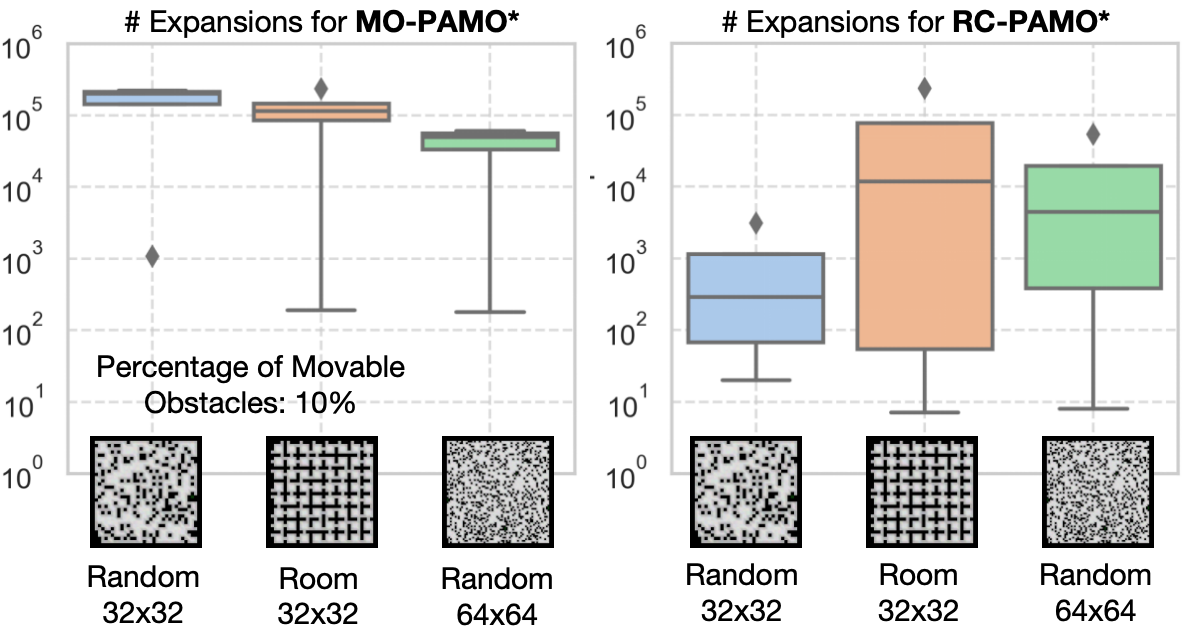}
\caption{Numbers of expansion of MO-PAMO* and RC-PAMO* in three grid maps Random 32x32, Room 32x32 and Random 64x64, with 10\% objects.
    }
\label{PAMO:fig:res_exp_random}
\end{figure}

\subsection{Results of PAMO*}

As shown in Fig.~\ref{PAMO:fig:res_rt_empty}, in the smaller Empty map, the increase in the number of movable obstacles slightly slows down the planning for both MO-PAMO* and RC-PAMO*.
In the larger maps with much more movable obstacles (Fig.~\ref{PAMO:fig:res_rt_random}), the planning is slowed down obviously, where MO-PAMO* times out in many instances before finding the entire Pareto-optimal front, while RC-PAMO* is still able to solve most of the instances.
This is expected since finding the entire Pareto-optimal front is usually much harder than finding a single optimal solution~\cite{ren22emoa,ren23erca}.

Fig.~\ref{PAMO:fig:res_exp_empty} and \ref{PAMO:fig:res_exp_random} shows the number of expansions of MO-PAMO* and RC-PAMO*.
Similar trends as the runtime can be observed for number of expansions.
Besides, we observe from Fig.~\ref{PAMO:fig:res_rt_random} and \ref{PAMO:fig:res_exp_random} that, in Random 32x32 and Random 64x64, as the map size and $|V_{mo}|$ increases, the runtime for RC-PAMO* increases while the number of expansion decreases.
The reason is, with a larger $|V_{mo}|$, each state has to encode the position of more movable obstacles and the processing time in each expansion increases correspondingly.

When considering the theoretic size of the entire state space $S$, the number of expansion is usually much smaller than the size of $S$.
For example, in Empty 8x8 with $|V_{mo}| = 12$, the size of the state space is $|S| = (8\times 8)^{(1+12)} \approx 10^{23}$ while the number of expansions is usually less than $10^3$ for MO-PAMO* and $10^2$ for RC-PAMO*.
It indicates that, guided by the heuristic, PAMO* only need to explore a small fraction of the state space to find optimal solutions.

\begin{figure}[tb]
\centering
\includegraphics[width=\linewidth]{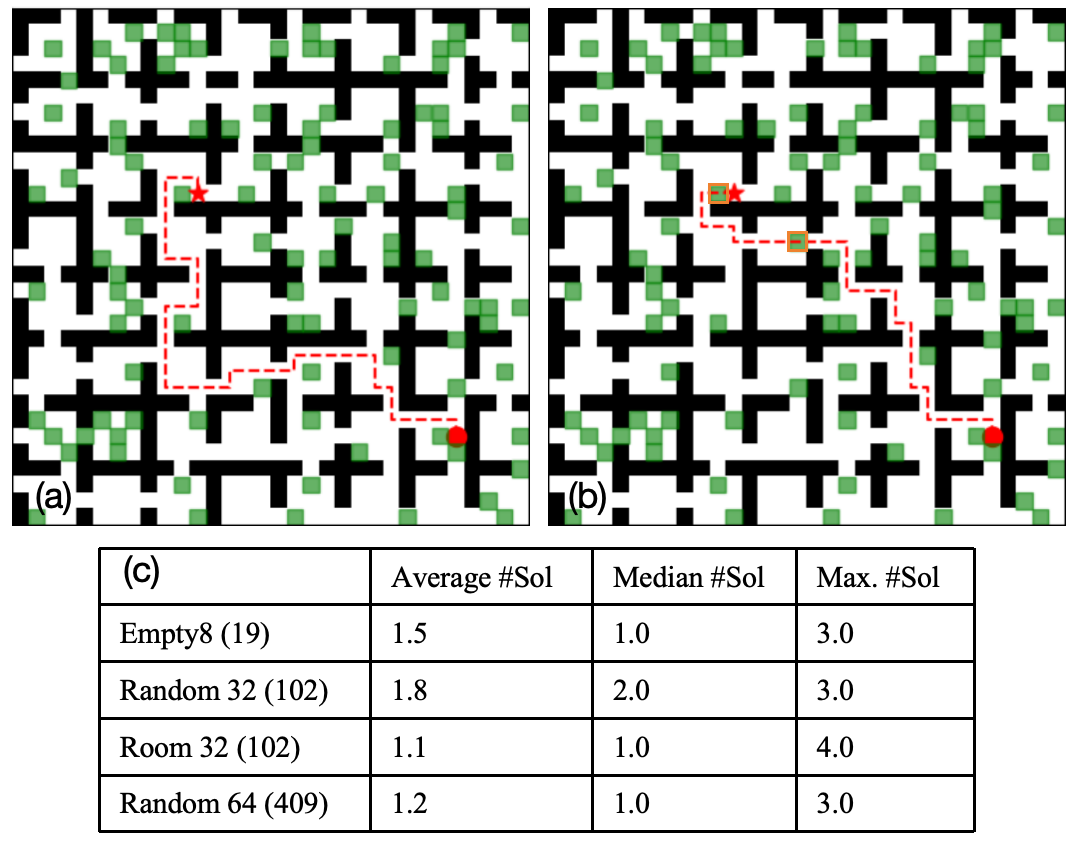}
\caption{Numbers of solutions found by MO-PAMO* within the runtime limit, and two snapshots of solution paths of an instance in Room 32x32.
    }
    \vspace{-4mm}
\label{PAMO:fig:res_num_sol}
\end{figure}

Finally, we show the number of solutions found by MO-PAMO* in Fig.~\ref{PAMO:fig:res_num_sol}.
Within the time limit, the planner finds 1 or 2 solutions as shown by the average and median number, while for some instances, the planner finds 3 or 4 solutions.
We pick an instance from Room 32x32 and shows the two different paths found by MO-PAMO*.
Fig.~\ref{PAMO:fig:res_num_sol}(a) shows a longer path with no push action required, while Fig.~\ref{PAMO:fig:res_num_sol}(b) shows a shorter path with 7 push actions.
These results can help determine which path to take based on the specific robotic platform where the push action is expensive or not.

\subsection{Results of H-PAMO*}

\begin{figure}[tb]
\centering
\includegraphics[width=\linewidth]{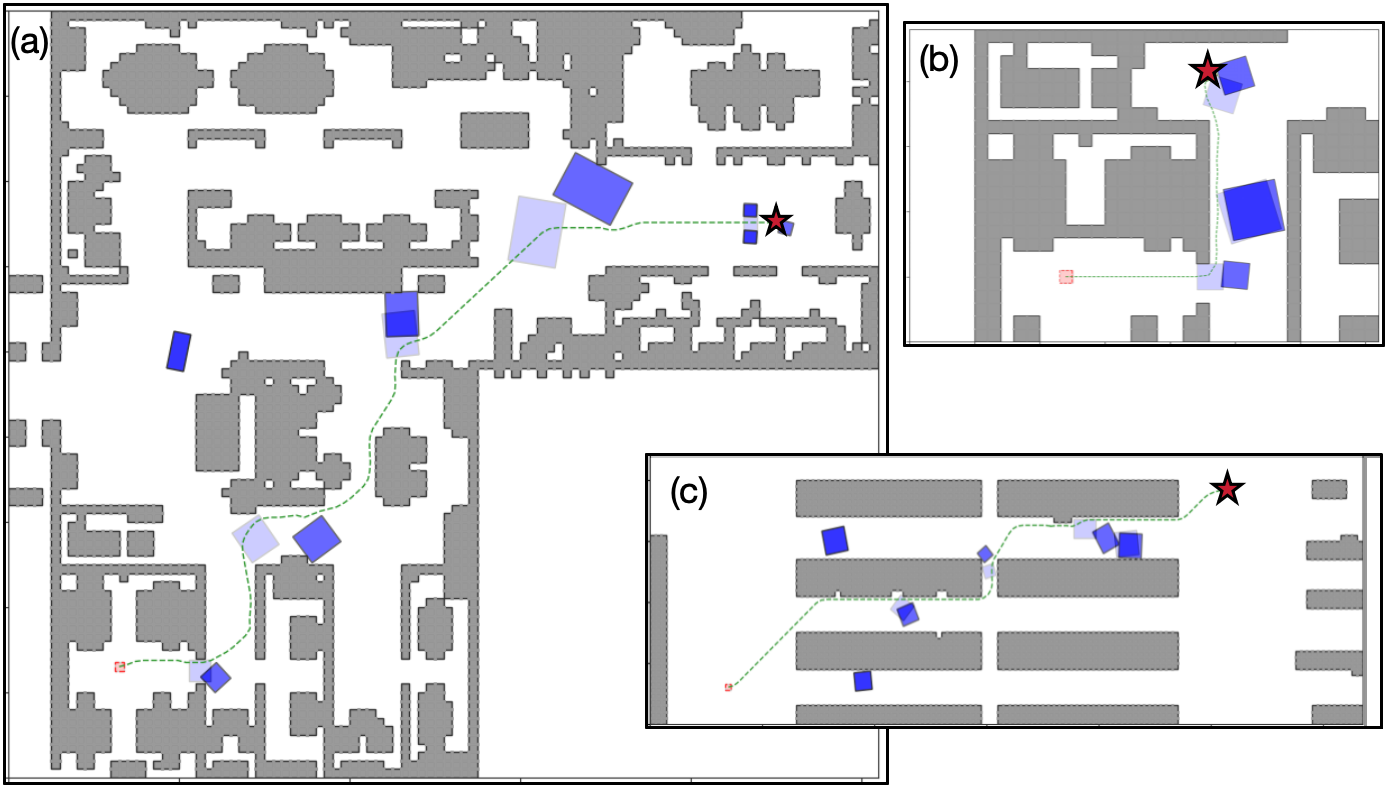}
\caption{Solution paths of H-PAMO* in three maps.
    }
\label{PAMO:fig:res_hpamo}
\end{figure}


\begin{figure}[tb]
\centering
\includegraphics[width=0.6\linewidth]{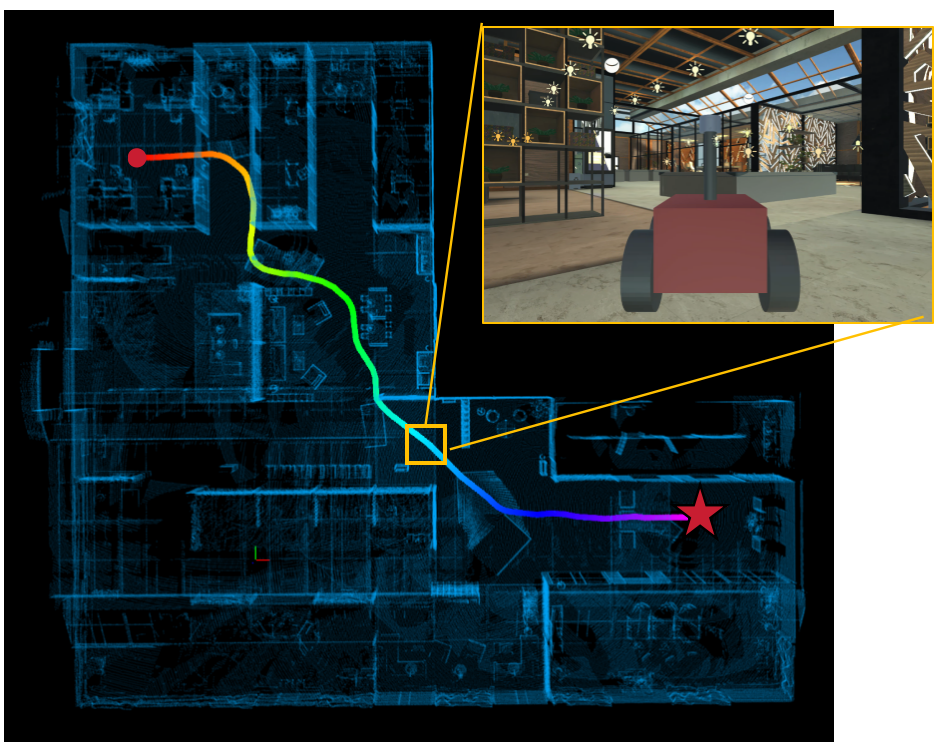}
\caption{Unity-Based Simulation of H-PAMO*.
    }
\label{PAMO:fig:res_sim}
\end{figure}

As shown in Fig.~\ref{PAMO:fig:res_hpamo}, we test H-PAMO* in three maps of sizes (a) 74x127 in an office, (b) 24x36 in part of an office, and (c) 90x102 in a warehouse, with rectangular objects of different sizes.
The robot is a 1x1 square.
The size of cells in $S_g$ for pruning is set to $(0.2,0.2,0.4)$ and the set of controls used are $(v,w)\in$ $\{(1.0,0.5)$, $(1.0,-0.5)$, $(1.0,0)$, $(-0.2,0)$, $(1.0,0.25)$, $(1.0,-0.25)$, $(0, 0.5)$, $(0, -0.5)\}$.
Fig.~\ref{PAMO:fig:res_hpamo} shows the solution paths, the initial and ending pose of all objects.
On all three instances, H-PAMO* terminates within 10 seconds and finds paths of length (a) 117.10 (b) 28.89 (c) 106.75 respectively.
Fig.~\ref{PAMO:fig:res_sim} shows the simulated motion of the robot for Fig.~\ref{PAMO:fig:res_hpamo}(a) in Unity3D.

%% file: conclude.tex
This paper investigates PAMO by formulating the problem as a multi-objective search and a resource constrained search over a grid, and develop planners with completeness and solution optimality guarantees.
The results verify that, although the state space is huge, in practice, only a small fraction of the state space needs to be explored during planning as guided by a heuristic, and most of the objects that are far from the robot's path are intact, which thus leads to algorithms that are often runtime efficient.
This paper also seeks to get rid of the grid representation by developing Hybrid-state PAMO* (H-PAMO*) to plan in continuous spaces at the cost of losing completeness and solution optimality guarantees.
One can consider the uncertainty of robot-object interaction, dynamic environments~\cite{ren22mopbd}, or multi-agent~\cite{ren22mocbs_tase} for future work.